\documentclass{article}
\usepackage{spconf,amsmath,graphicx}
\usepackage{xcolor}
\usepackage{subcaption}
\usepackage{tabularx}
\usepackage{tikz}
\usepackage{amssymb}
\usepackage{verbatim,makecell}

\def\vblank{\langle\textit{\ignorespaces blank \unskip}\rangle}

\def\checkmark{\tikz\fill[scale=0.4](0,.35) -- (.25,0) -- (1,.7) -- (.25,.15) -- cycle;}

\definecolor{darkgreen}{HTML}{006400}
\definecolor{darkyellow}{HTML}{cccc00}

\title{Language Model fusion for streaming end to end speech recognition}
\makeatletter
\def\@name{\emph{Rodrigo Cabrera}, \emph{Xiaofeng Liu}\sthanks{Performed the work while at Google}, \emph{Mohammadreza Ghodsi}, \\
\emph{Zebulun Matteson}, \emph{Eugene Weinstein}, \emph{Anjuli Kannan}}
\makeatother
\address{
\texttt{\small\{rodrigocabrera,ghodsi,zmatteson,weinstein, anjuli\}@google.com, xiaofeng.liu@twosigma.com} \\
Google Inc.}
\begin{document}
\maketitle

\begin{abstract}
Streaming processing of speech audio is required for many contemporary
practical speech recognition tasks. Even with the large corpora of
manually transcribed speech data available today, it is impossible for
such corpora to cover adequately the long tail of linguistic content
that's important for tasks such as open-ended dictation and voice
search. We seek to address both the streaming and the tail
recognition challenges by using a language model (LM) trained on
unpaired text data to enhance the end-to-end (E2E) model.
We extend shallow fusion and cold fusion approaches to
streaming Recurrent Neural Network Transducer (RNNT), and also propose
two new competitive fusion approaches that further enhance the RNNT architecture.
Our results on multiple languages with varying training set sizes show that
these fusion methods improve streaming RNNT performance through introducing
extra linguistic features. Cold fusion works consistently better on streaming
RNNT with up to a \textbf{8.5\%} WER improvement.

\end{abstract}


\section{Introduction}
\label{sec:intro}

End-to-end (E2E) models for automatic speech recognition (ASR) tasks have gained
popularity because these models predict subword sequences from acoustic features
with a single model, unlike \textit{classic ASR} systems which have separate
acoustic, pronunciation, and language model components. The most common E2E
architectures are either attention-based (e.g., listen, attention, and spell
(LAS)~\cite{chan_las_icassp2016}) or RNNT~\cite{graves_arxiv2012}
models (see~\cite{prabhavalkar_is2017} for comparison).

Overall, E2E models show comparable or better word error rate (WER) performance
with a simplified system setup. While standard attention-based models
must inspect the entire input sequence before generating outputs,
streaming-friendly modifications of such models have been proposed,
such as MoChA~\cite{Mocha17} and neural transducer~\cite{NeuralTransducer16}.

While these approaches have shown promise, the RNNT architecture is an
alternative E2E model that can natively predict output sequences on
the fly (with unidirectional encoders), and thus is a natural choice
for streaming applications. RNNT models have also shown extremely strong
results., e.g., He et al.~\cite{he_icassp2019}
presented an RNNT based real-time steaming recognizer, which outperformed classic
ASR models by a wide margin.

Classic ASR models leverage unpaired text data with a separately trained
language model (LM) and second-pass rescoring model~\cite{biadsy_is2017},
but unpaired text data cannot be easily utilized when training E2E
models. Although E2E models have overall shown strong results, they
have been shown to have difficulty accurately modeling tail
phenomena such as proper nouns, numerics, and accented
speech~\cite{Crosslingual1998,NonNative2014,Numeric2019,ProperNoun2020},
due to the requirement that they be trained on paired (speech-transcript) data.

Recent papers have proposed fusing E2E models with LMs trained with text data (usually referred
to this as \textit{fusion}), including shallow fusion~\cite{glehre_2015OnUM,kannan_icassp2018},
deep fusion~\cite{glehre_2015OnUM}, cold fusion~\cite{sriram_archiv2015}, component
fusion~\cite{shan_icassp2019}, etc. Most experiments used neural LMs, and some used
n-gram fst LMs~\cite{chan_las_icassp2016,bahdanau_icassp2016,glehre_2015OnUM,chorowski_is2017}.
See~\cite{Toshniwal2018ACO} for comparison of some of these approaches.
However, these experiments were performed with standard non-streaming
attention models. Fusion approaches with streaming E2E models have been
unexplored, and it is unknown whether the observe gains on attention-based models
could be translated to streaming models.

In this paper, we explore shallow fusion, cold fusion and two new fusion approaches
unique to the streaming RNNT models in section \ref{sec:methods}, we detail the experimentation
across multiple languages of varying sizes of training data in section \ref{sec:setup},
and we analyze the results in section \ref{sec:results}. We show that
while shallow fusion worked better than cold fusion for attention-based
models, cold fusion outperforms shallow fusion for
RNNT models, with a WER reduction of up to \textbf{8.5\%}.

\section{Methods}
\label{sec:methods}

\subsection{RNN Transducer}
\label{sec:rnnt}
The RNNT architecture proposed by Graves~\cite{graves_arxiv2012} consists of
an encoder, a prediction network, and typically a joint network.
It directly predicts a sequence of words, subwords~\cite{Schuster2012} or graphemes
without using an external pronunciation model or language model.
Like Connectionist temporal classification (CTC), its target symbol
set is augmented with a blank symbol ($\vblank$) such
that the RNNT model does not output a grapheme or subword symbol at every time frame.
The RNNT outputs a symbol at each time frame $t = 1, 2, \ldots, T$ where $T$ is the total number of
frames, the input of the encoder is the log-mel filterbank energies of
dimension $d$.
The input of the prediction network is the last non-$\vblank$ symbol during
prediction.
The joint network takes as input the output vectors of both encoder and
prediction networks and outputs logits which are then passed to the softmax
layer to predict subword symbols.
In this work, we focus on streaming RNNT wordpiece models, where the encoder and
prediction networks both use unidirectional LSTM layers, and the predicted symbols
are wordpieces.
During decoding, a blank penalty is usually added to adjust the posterior
probability of the $\vblank$ symbol, and its value is usually optimized through
parameter sweep.

\subsection{Fusion with RNN-LM}
The recurrent neural network (RNN) based LM predicts
the probability of a symbol given the context, and is trained with text-only
data. It contains an embedding layer followed by a stack of
unidirectional RNN layers.
For a sequence of wordpieces ${w}_{1}^{T}$, the RNN-LM computes a probability:
\begin{equation}
P({w}_{1}^{T}) = \prod_{i=1}^{T} P(w_i | w_1, w_2, \ldots, w_{i-1}).
\end{equation}

One big difference between attention-based models and RNNT models
is the existence of the $\vblank$ symbol in RNNT models, and that symbols needs to be
handled correctly during fusion.
While the RNN-LMs are trained without knowledge of the $\vblank$ symbol, the LM probability
of $\vblank$ must be defined during inference time.
In this work we make the language model's probability of $\vblank$ be equal to the RNNT
probability,i.e.,$\log P_{lm}(\vblank) = \log P_{rnnt}(\vblank)$,
where $\log P_{lm}$ and
$\log P_{rnnt}$ are the probability of LM and RNNT, respectively.
This means that when the RNNT model outputs $\vblank$, the RNN-LM is not
updated and that the probability for the $\vblank$ symbol after fusion
remains unchanged.

\subsection{Shallow Fusion}
\begin{figure}[t]
  \centering
  \begin{subfigure}[b]{0.23\textwidth}  
    \centering
    \includegraphics[width=\textwidth]{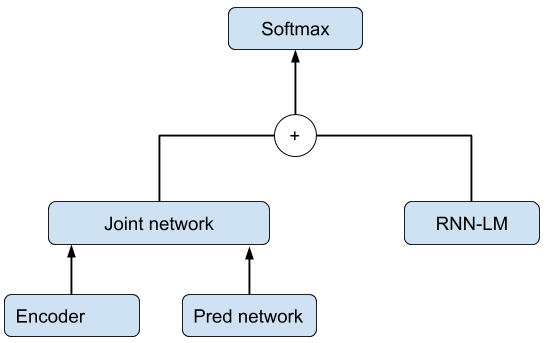}
    \caption{Shallow fusion}
    \label{fig:shallow_fusion}
  \end{subfigure}
  \hfill
  \begin{subfigure}[b]{0.23\textwidth}  
    \centering
    \includegraphics[width=\textwidth]{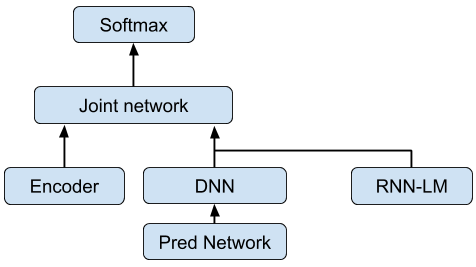}
    \caption{Early shallow fusion}
    \label{fig:early_shallow_fusion}
  \end{subfigure}
  \begin{subfigure}[b]{0.23\textwidth}  
    \centering
    \includegraphics[width=\textwidth]{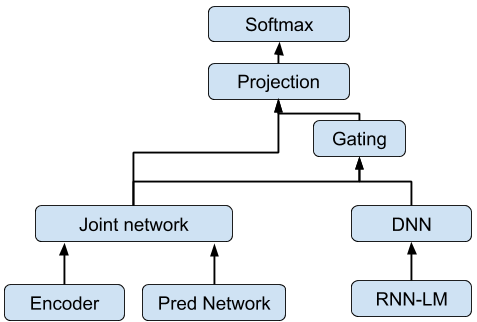}
    \caption{Cold fusion}
    \label{fig:cold_fusion}
  \end{subfigure}
  \hfill
  \begin{subfigure}[b]{0.23\textwidth}  
    \centering
    \includegraphics[width=\textwidth]{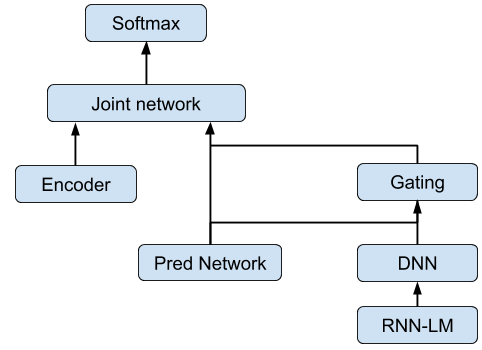}
    \caption{Early cold fusion}
    \label{fig:early_cold_fusion}
  \end{subfigure}
  \caption{{Illustrations of our different fusion styles.}}
  \vspace{-0.2in}
\end{figure}

Shallow fusion (SF), first proposed by Gulchere, et al.~\cite{glehre_2015OnUM},
integrates an external LM with the RNNT only during inference time, as shown in
Fig~\ref{fig:shallow_fusion}.
The scores of RNNT and RNN-LM are log-linearly combined before the softmax
layer.
\begin{equation}
  \begin{aligned}
  \begin{gathered}
    \log P(y_t) = \log P_{rnnt}(y_t) + \beta \log P_{lm}(y_t),
  \end{gathered}
  \end{aligned}
  \label{eq:shallow_fusion}
\end{equation}
where $\beta$ is the LM weight, and $y_t$ is the output symbol at time frame
$t$.

In SF, the RNNT and the LM are trained independently with
different training data, and thus the approach is very modular and it is easy
to integrate the LM during inference time. Note that shallow fusion performance is
sensitive to the LM weight $\beta$ and to the RNNT's blank penalty,
and these values usually need to be optimized through sweeping when using the LM.

\subsection{Early Shallow Fusion}

One popular hypothesis~\cite{prabhavalkar_is2017} about the RNNT architecture
is that the prediction network is analogous to an LM.
Inspired by this hypothesis we propose the early shallow fusion (ESF) approach
as illustrated in Fig.~\ref{fig:early_shallow_fusion}
It fuses the outputs of the pretrained LM with the RNNT prediction network
using log-linear interpolation before feeding to the joint network, i.e.,
\begin{equation}
  \begin{aligned}
  \begin{gathered}
    l_t^{Pred} = DNN(h_t^{Pred})\\
    h_t^{ESF} = l_t^{Pred} + \beta l_t^{LM}
  \end{gathered}
  \end{aligned}
  \label{eq:early_shallow_fusion}
\end{equation}
Where $h_t^{Pred}$ is projected to have a logit output and $l_t^{LM}$ is the logit output
of the external LM at time frame $t$, $\beta$ is the LM weight,
and $h_t^{ESF}$ is the interpolated vector that is fed to the joint network.
Similar to SF, the LM is used only during inference time and not for RNNT
training. It's possible to fine tune the RNNT model by
further training it for extra steps with the fused LM on supervised data.

\subsection{Cold Fusion}

In cold fusion (CF)~\cite{sriram2017cold}, the model is trained using
a pre-trained LM that remains fixed when training the rest of the model.
A fine-grained gating approach is applied on the LM
logit instead of the hidden state to allow for flexible swapping and to improve
performance on out-of-domain scenarios as in component fusion~\cite{shan_icassp2019}.
The model architecture is illustrated in Fig.~\ref{fig:cold_fusion}, defined as:

\begin{equation}
  \begin{aligned}
  \begin{gathered}
    h_t^{LM} = DNN(l_t^{LM})\\
    g_t = \sigma(W[s_t^{jn};h_t^{LM}] + b)\\
    s_t^{CF} = [s_t^{jn};g_t \dot h_t^{LM}]\\
    r_t^{CF} = DNN(s_t^{CF})\\
    \hat{P}(y_t|x,y_{<t}) = softmax(r_t^{CF})
  \end{gathered}
  \end{aligned}
\label{eq:cold_fusion}
\end{equation}
Where $l_t^{LM}$ is the logit output of the language model, $g_t$ is the fine-grained gate output parametrized by $W$,
which controls the importance of the contribution of the hidden state of the LM.
$s_t^{jn}$ is the state of the joint network, $s_t^{CF}$ is the final fused state,
and $DNN$ is a feedforward neural network. In our experiments we have been using
a single layer of fully connected network.

Training the $DNN$ and sigmoid functions help the model learn to use the LM when predicting
a non-$\vblank$ symbol, and only use the RNNT when predicting the $\vblank$ symbol.
In the CF approach, sweeping for LM weight and blank penalty is not needed.

\subsection{Early Cold Fusion}
We propose the early cold fusion (ECF) approach as illustrated in
Fig.~\ref{fig:early_cold_fusion}. ECF is inspired by cold fusion and
fuses the LM by passing its logits through a $DNN$ layer, the $DNN$ output is
concatenated with the hidden states of the prediction network and fed into a gating layer
before sending to the joint network.
The LM is used during training and inference.

\begin{equation}
  \begin{aligned}
  \begin{gathered}
    h_t^{LM} = DNN(l_t^{LM})\\
    g_t = \sigma(W[h_t^{Pred};h_t^{LM}] + b)\\
    h_t^{ECF} = [h_t^{Pred};g_t \dot h_t^{LM}]\\
  \end{gathered}
  \end{aligned}
  \label{eq:early_cold_fusion}
\end{equation}

\section{Experiment Setup}
\label{sec:setup}

\subsection{Model Architecture}
The RNNT models and LMs in all experiments shared the same architectures.
For RNNT, the encoder network consists of 8 unidirectional LSTM layers, where each
layer has 2048 hidden units followed by a 640-dimensional projection layer.
A stacking layer is inserted after the second layer, which stacks the outputs
of two neighboring frames with a stride of 2 for speed improvement.
The prediction network contains 2 unidirectional LSTM layers, where each layer
has 2048 hidden units and followed by a 640-dimensional projection layer.
The outputs of the encoder and prediction networks were fed into the joint
network.
The joint network is a feed-forward network with 640 hidden units, which
accepts input from both the encoder and the prediction networks.
The overall RNNT model has 120 million parameters.

The RNN-LM model has 2 unidirectional LSTM layers, each with 2048 units, with
an embedding layer of 128 dimensions. This model has 60 million parameters.
Both the RNNT and LM models were trained with using wordpieces and a vocabulary size
of 4096.

\subsection{Data}
We performed experiments on three languages: Greek, Norwegian, and Sinhala.
The training sets consisted of ~6,700, 3,500, and 160 hours of utterances
respectively, which were anonymized and hand-transcribed, and are
representative of Google's voice search traffic.
These languages covered the cases with low to high amount of training data.
During training, the utterances were artificially noisified using a room
simulator; the noise and reverberation had an average signal-to-noise
ratio of 12dB \cite{Chanwoo17}. The noise sources came from YouTube or noisy
environmental recordings.

For the RNN-LM, during training the data were randomly drawn from a mix of
text sources with different weights: 0.6 for the same hand-transcribed data used
in RNNT training, and 0.1 for each of YouTube search logs, Google search
queries, Maps search queries and crawled web documents.
This in total accounted for over 200 million text sentences for Sinhala
and over a billion sentences for Greek and Norwegian.
The RNN-LMs are trained by minimizing the log perplexity over a held out set.
For all the languages, the fused models were tested on anonymized voice search testsets,
which were drawn from Google's speech traffic and were not used while training the model.
For shallow fusion models, the LM weight and blank penalty were also optimized
by sweeping using these testsets.

\section{Results}
\label{sec:results}

\begin{table}[th]
  \caption{Log perplexity of the LMs over the VS testset.}
  \label{tab:perplexity}.
  \begin{tabular}{|p{0.4\linewidth}|p{0.4\linewidth}|}
    \hline
     & Log Perplexity\\
    \hline
    Norwegian & 6.64\\
    Greek &  6.75\\
    Sinhala & 11.92\\
    \hline
  \end{tabular}
  \vspace{-0.1in}
\end{table}

For each of the three languages, the LM was trained once with
the text-only data, and applied in all the experiments.
Table~\ref{tab:perplexity} shows the perplexities of the pretrained LMs when
predicting the test sets.
It can be seen that it was much higher for Sinhala than Greek
or Norwegian. This is expected because the Sinhala LM training data were
dominated by the non-transcribed text sources which might be more different from
the test sets.

\subsection{Shallow fusion and cold fusion}

\begin{table}[th]  
  \caption{Shallow fusion and cold fusion results.}
  \label{tab:rnnt_sf_cf}
  \begin{tabular}[t]{|p{0.12\columnwidth}|p{0.12\columnwidth}|p{0.18\columnwidth}|p{0.15\columnwidth}|p{0.16\columnwidth}|}
  \hline
    Model & \#params & Norweigian & Greek & Sinhala \\
    \hline
    Baseline & 120M &
    \makecell[c]{24.5} & \makecell[c]{14.5} & \makecell[c]{69.4} \\
    \hline
    LPN & 191M &
    \makecell[c]{24.4 \\(\textcolor{darkgreen}{-0.40\%})} &
    \makecell[c]{\textit{N/A}} &
    \makecell[c]{70.0 \\(\textcolor{red}{+0.86\%})} \\
    \hline
    SF & 180M &
    \makecell[c]{24.2 \\(\textcolor{darkgreen}{-1.22\%})} &
    \makecell[c]{14.4 \\(\textcolor{darkgreen}{-0.68\%})} &
    \makecell[c]{69.3 \\(\textcolor{darkgreen}{-0.14\%})} \\
    \hline
    CF & 200M &
    \makecell[c]{22.4 \\(\textcolor{darkgreen}{-8.57\%})} &
    \makecell[c]{14.0 \\(\textcolor{darkgreen}{-3.44\%})} &
    \makecell[c]{69.0 \\(\textcolor{darkgreen}{-0.57\%})} \\
    \hline
  \end{tabular}
  \vspace{-0.1in}
\end{table}

Table~\ref{tab:rnnt_sf_cf} shows the performance in terms of WER on the testsets
for the baseline model, a modified RNNT model with larger prediction
network (LPN), shallow fusion (SF), and cold fusion (CF).
The LPN was identical to the baseline model except its prediction network
had 6 LSTM layers instead of 2, such that the number of parameters was similar
to the CF model.
The LPN model on Norweigian showed slight improvement over the baseline
model, suggesting that just increasing RNNT model size is not effective for
lower resource languages.
Overall, cold fusion performed significantly better than shallow fusion.
This is surprising as it's the opposite pattern than what was seen
with LAS models~\cite{Toshniwal2018ACO}.
Shallow fusion gave a small gain (1.2\%) for Norweigian, but had no significant
improvement for Greek and Sinhala, while cold fusion improved on all three languages.
This could be because the handling of $\vblank$ probability in SF
was not ideal, while CF can benefit from the
extra parameters and the co-training of the RNNT model with the LM where the
RNNT model learned to adapt to the fused LM.

Notably for Norwegian, cold fusion showed 8.57\% gains, which was much
larger than that for Greek (3.44\%).
This could be because the RNNT model for Greek was trained with a much larger
data set than Norwegian. The large train set covers more
linguistic features compared to the coverage in Norwegian, and as a result the gain brought
by extra text-data was smaller.
For Sinhala, both methods didn't show significant improvement. Our hypothesis is that the training
set for Sinhala was too small, causing the model to overfit quickly during training;
and that the linguistic features of the trained LM are too different from the
test sets as manifested in the LM perplexity shown in Table~\ref{tab:perplexity}.
These results impliy that the amount of training data is critical for RNNT
performance.

\begin{table}
  \caption{Examples of the top hypothesis generated by the baseline and the cold fused RNNT models for Greek.}
  \label{tab:hyp_compare}
  \begin{tabular}[t]{|p{0.16\linewidth}|p{0.32\linewidth}|p{0.32\linewidth}|}
    \hline
    & Example 1 & Example 2 \\
    \hline
    \makecell[tl]{Baseline} &
    \makecell[l]{philakes \textbf{\color{red}ipsous tis}\\ \textbf{\color{red}Afstralias}\\andrianoupolis} &
    \makecell[l]{\textbf{\color{red}etsi re spor} pou\\ anevainis skalopatia} \\
    \hline
    \makecell[tl]{Cold fused\\RNNT} &
    \makecell[l]{philakes \textbf{\color{darkgreen}ipsistis}\\ \textbf{\color{darkgreen}asphalias} (\checkmark) \\ andrianoupolis} &
    \makecell[l]{\textbf{\color{darkgreen}Range Rover sport} \\(\checkmark) pou anevainis\\skalopatia} \\
    \hline
    \makecell[tl]{English\\Translation} &
    \makecell[tl]{\textbf{\color{darkgreen}highest security}\\(\checkmark) prisons in \\Andrianopolis} &
    \makecell[tl]{\textbf{\color{darkgreen}Range Rover sport} \\(\checkmark) that climbs \\stairs} \\
    \hline
  \end{tabular}
  \vspace{-0.1in}
\end{table}

Table~\ref{tab:hyp_compare} shows two examples of the top hypothesis generated
by the baseline RNNT model and cold fusion in Greek. The hypotheses were
transliterated to Latin for better understanding, and the differences between
the hypotheses of the two models were highlighted.
The last row shows the English translations of the transcript truth.
The fused model correctly recognized the utterances while the baseline model
failed to produce hypotheses that makes sense.
This was likely because these mistaken words were seen in the text-only data
used in LM training while not in the transcribed speech-text data used
in the RNNT training. This example illustrates how fusion cat help a streaming E2E
model perform better on ``tail'' phenomena such as proper nouns.
Fusion with the LM helped the RNNT model to obtain a richer
lattice and surface better hypotheses, and thus resulted in correct inference.

\subsection{Early shallow fusion and early cold fusion}

\begin{table}
  \caption{Early shallow fusion and early cold fusion results.}
  \label{tab:early_rnnt_wer}
  \begin{tabular}[t]{|p{0.12\columnwidth}|p{0.12\columnwidth}|p{0.18\columnwidth}|p{0.15\columnwidth}|p{0.15\columnwidth}|}
    \hline
    Model & \#params & Norweigian & Greek & Sinhala \\
    \hline
    Baseline & 120M & 24.5 & 14.5 & 69.4 \\
    \hline
    ESF & 180M &
    \makecell[c]{24.4 \\(\textcolor{darkgreen}{-0.40\%})} &
    \makecell[c]{14.5 \\(\textcolor{darkyellow}{0.00\%})} &
    \makecell[c]{69.3 \\(\textcolor{darkgreen}{-0.14\%})} \\
    \hline
    ECF & 200M &
    \makecell[c]{22.7 \\(\textcolor{darkgreen}{-7.34\%})} &
    \makecell[c]{14.2 \\ (\textcolor{darkgreen}{-2.06\%})} &
    \makecell[c]{68.1 \\(\textcolor{darkgreen}{-1.87\%})} \\
    \hline
  \end{tabular}
  \vspace{-0.1in}
\end{table}

Table~\ref{tab:early_rnnt_wer} shows the results of early shallow fusion and
early cold fusion, where the LMs were fused before the joint network. It's worth
noting that ESF models needed fine-tuning to get improvements.
Similar to the results in Table~\ref{tab:rnnt_sf_cf}, Sinhala
had the least WER gain while Norweigian had the most gain because of the same
reasons.
For Norweigian and Greek, both early fusion approaches showed significant
gains in WER, but slightly less than SF and CF.
This suggests that fusing the LM before or after the joint network
produces similar results. This result is valuable in that it shows us
that if we choose to modify the topology of the RNNT
model~\cite{StatelessPrediction2020} we can be flexible with the
placement of the fusion point while maintaining the superior quality
afforded by fusing a text-trained LM.

\section{Discussion}
\label{sec:discussion}

In this work we explored several existing fusion approaches as well as new methods
that fuse a pre-trained language model at an earlier stage to
improve RNNT performance for medium and low resource languages in a streaming setting.
Among all the approaches, cold fusion performed best, with WER reduction up to
\textbf{8.5\%} compared to the baseline. The language model brings
additional linguistic features and helps the RNNT to
produce richer lattices and obtain better hypotheses. Additionally, we showed that
the fusion point can be placed either before or after the joint
network while maintaining the quality gains. 
\vfill\pagebreak

\bibliographystyle{IEEEbib}
\bibliography{refs}

\end{document}